\documentclass{INTERSPEECH2023}
\usepackage{fdsymbol}

\interspeechcameraready 

\title{Enhancing Black-Box Few-Shot Text Classification with Prompt-Based Data Augmentation}

\name{Danqing Luo$^\dag$, Chen Zhang$^\dag$, Jiahui Xu$^\dag$ \\
Bin Wang$^\star$, Yiming Chen$^\dag$, Yan Zhang$^\dag$, Haizhou Li$^{\varheartsuit,\dag,\diamondsuit}$}
\address{
  $^\dag$National University of Singapore  $^\star$Institute of Infocomm Research (I2R), A*STAR, Singapore\\
  $^\varheartsuit$The Chinese University of Hong Kong, Shenzhen, China $^\diamondsuit$Kriston AI Lab, China}
\email{danqing@nus.edu.sg}

\begin{document}

\maketitle
 
\begin{abstract}
Training or finetuning large-scale language models (LLMs) such as GPT-3 requires substantial computation resources, motivating recent efforts to explore  parameter-efficient adaptation to downstream tasks. One practical area of research is to treat these models as black boxes and interact with them through their inference APIs. In this paper, we investigate how to optimize few-shot text classification without accessing the gradients of the LLMs. To achieve this, we treat the black-box model as a feature extractor and train a classifier with the augmented text data. Data augmentation is performed using prompt-based finetuning on an auxiliary language model with a much smaller parameter size than the black-box model. Through extensive experiments on eight text classification datasets, we show that our approach, dubbed BT-Classifier\footnote{Black-box Text Classifier}, significantly outperforms state-of-the-art black-box few-shot learners and performs on par with methods that rely on full-model tuning.

\end{abstract}

\section{Introduction}

In the past few years, significant progress has been made in research on large-scale language models (LLMs)~\cite{devlin-etal-2019-bert,liu2019roberta}. Scaling up language models has been demonstrated to boost performance and sample efficiency on a great variety of downstream tasks~\cite[\textit{inter alia}]{raffel-etal-2020-exploring,brown-etal-2020-language}. However, training such LLMs is not practical with typical research hardware. Even finetuning them on task-specific data is extremely challenging. Many research efforts have been devoted to more parameter-efficient adaptation approaches, including (1) parameter-efficient tuning (PET)~\cite{lester-etal-2021-power,li-liang-2021-prefix,houlsby-etal-2019-parameter}, which optimizes a small portion of task-specific parameters, while keeping the language model intact; (2) In-context learning (ICL)~\cite{brown-etal-2020-language}, which requires no parameter tuning but relies on input-output demonstrations specific to the task at hand; (3) derivative-free optimization (DFO)~\cite{sun-etal-2022-bbt,sun-etal-2022-bbtv2}, which injects task-specific prompt tokens into the input sequence and adopts derivative-free optimization methods, such as evolution algorithms~\cite{hansen-ostermeier-2001-completely}, for continuous prompt optimization. 


Although the adaptation approaches mentioned above are on par with full-model fine-tuning in terms of performance and are more efficient in terms of parameters, they still face several limitations. PET methods, such as adapters~\cite{houlsby-etal-2019-parameter} and continuous prompt tuning~\cite{lester-etal-2021-power,li-liang-2021-prefix}, still require access to the gradients and architecture of the LLMs, i.e., they need to perform backpropagation through the entire LLM. Consequently, the computation costs remain high and the LLMs need to be transparent. Moreover, ICL is highly sensitive to input example selection and input template design~\cite{gao-etal-2021-making,zhao-etal-2021-calibrate}. Its performance is also unstable across different downstream tasks. Hence, it is impractical to deploy ICL for real-world use. Lastly, the optimization process of DFO methods can be quite slow, requiring tens of thousands of forward passes through the LLMs to achieve satisfactory performance for a small training data size. Additionally, these methods are prone to overfitting in the few-shot setting, and their slow optimization process makes it challenging to overcome this issue via data augmentation.


In this paper, we aim to enhance few-shot text classification with the power of LLMs. Yet, due to hardware constraints and the inaccessibility of most LLMs, we propose to conduct parameter-efficient adaptation of LLMs with a simple multi-layer perceptron (MLP) leveraging the inference APIs of LLMs. More specifically, we treat the black-box LLM as a feature extractor. Hidden states w.r.t. input text sequences are obtained via the inference APIs. An MLP classifier is trained on the hidden states and the corresponding labels. Despite its simplicity, the approach does not face the above-mentioned limitations. We name our approach BT-classifier.
First, unlike PET, BT-classifier does not need to backpropagate through the LLM during training, making it compatible with LLM inference APIs.
Second, as long as there are sufficient labeled data, the performance of BT-classifier is not sensitive to the input text sequences and stable across different classification tasks. Lastly, the training process can be quite fast.

\begin{figure}[t]
  \centering
  \includegraphics[width=\linewidth]{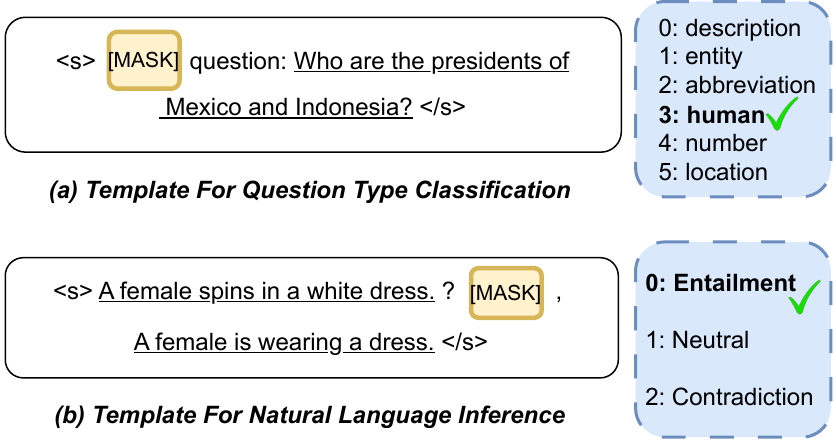}
  \caption{Input template examples. The blue boxes contain the labels for the corresponding classification tasks. The hidden states of the ``[MASK]" token extracted from the large-scale language model are used for training BT-Classifier.}
  \label{fig:input-example}
\end{figure}

A major challenge of BT-classifier for few-shot text classification is the lack of a sufficient amount of labeled data. In a typical few-shot setting, the number of labeled examples per class is less than 50. Depending on the task, hundreds or even thousands of labeled examples are necessary for the classifier to achieve satisfactory performance. To tackle this challenge, we propose a data augmentation technique that leverages prompt-based learning on an auxiliary language model, which is significantly  smaller than the black-box language model (discussed in \S\ref{subsec:data-augmentation}). Additionally, to fully exploit the semantic representation capability of the black-box LLM and align with the mask language modeling (MLM) pretraining objective\footnote{In our paper, we experiment with RoBERT-Large~\cite{liu2019roberta}, a large-scale pretrained language model based on transformer encoder. It serves as the backbone language model for many approaches for few-shot text classification tasks.}, the hidden states w.r.t. the ``[MASK]" token extracted from black-box LLM are used as input to the MLP classifier. 

In summary, our contributions are two-fold. First, we introduce BT-classifier, a fast and memory-efficient pipeline for adapting LLMs to the downstream few-shot text classification tasks. The entire pipeline, which includes the data augmentation process and training of the MLP classifier, can be completed on typical research hardware, such as a single 11GB 2080 TI GPU card. In addition, BT-Classifier is model-agnostic in the sense that it can be applied to any large-scale language model. Secondly, through extensive experiments on 8 text classification datasets, we demonstrate that BT-Classifier achieves state-of-the-art performance without tuning any parameter of the black-box language model.  

\section{Method}

\begin{figure}[!t]
  \centering
  \includegraphics[width=\linewidth]{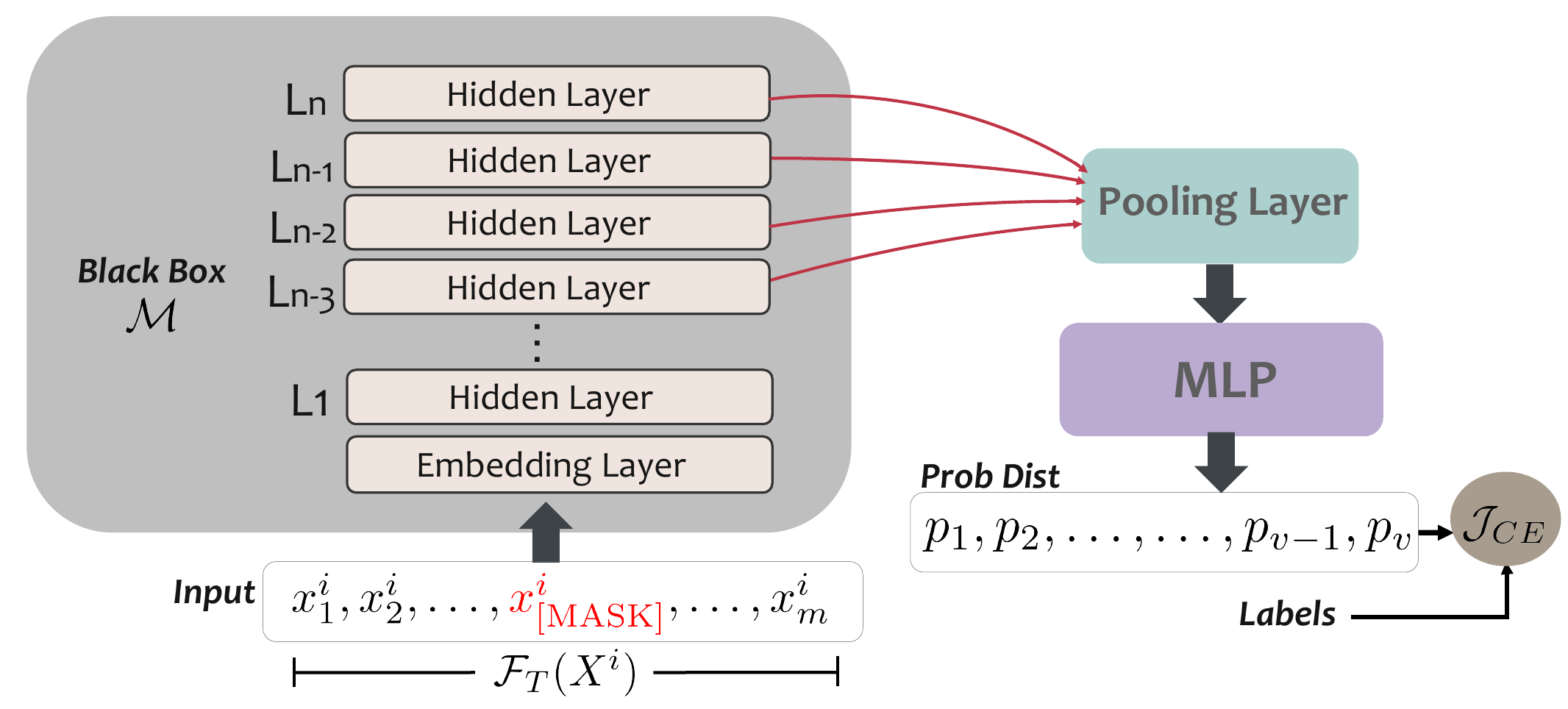}
  \caption{Overview of BT-Classifier}
  \label{fig:framwork}
\end{figure}

\subsection{Task Formulation}
\label{subsec:task-formulation}

In a few-shot text classification task $T$ with a label space $\mathcal{Y}$, we assume there are $K$ labeled training examples per class in the training set, $\mathcal{D}^T_{train}$. The training data size, $|\mathcal{D}^T_{train}| = K \times |\mathcal{Y}|$. We also assume an development set, $\mathcal{D}^T_{dev}$, which is of equal data size as $\mathcal{D}^T_{train}$. Both $\mathcal{D}^T_{train}$ and $\mathcal{D}^T_{dev}$ consist of data instances $(X^i, y^i)$ where $y^i\in{\mathcal{Y}}$ and $X^i$ denotes the input text sequence, which contains $n$ tokens, i.e., $X^i = \{x^i_1, x^i_2, \ldots, x^i_n\}$. Assume that we have task-specific template mapping function $\mathcal{F}_T$, which maps $X^i$ to a specific input format $\mathcal{F}_T(X^i)$. 
Figure~\ref{fig:input-example} shows two examples of $\mathcal{F}_T(X^i)$.
The underlined texts in the boxes are the original input texts, $X^i$. Moreover, assume a black-box LLM denoted as $\mathcal{M}$, which is for inference only. Through its cloud-based API, we can obtain the logits of ``[MASK]" tokens and the hidden states of the input text sequences. 
Our goal is to develop a model that generalizes well to an unseen test set $\mathcal{D}^T_{test}$. 

\subsection{Details of BT-Classifier}
\label{subsec:bt-overview}

Figure~\ref{fig:framwork} presents the overall architecture of BT-classifier. $\mathcal{M}$ serves as a black-box feature extractor and we can obtain the hidden states of the transformer layers via an inference API. When feeding $\mathcal{F}_T(X^i)$ into $\mathcal{M}$, we obtain a sequence of hidden vectors after each layer $l$. As we are interested in the hidden vectors w.r.t. the ``[MASK]" token in $\mathcal{F}_T(X^i)$ that is $\{\textbf{h}^{i,l}_{\textit{[MASK]}} \in \mathbb{R}^d \}_{l=1}^{L}$, we perform max pooling on $\{\textbf{h}^{i,l}_{\textit{[MASK]}}\}_{l=L-3}^L$ to derive a single vector representation, $\textbf{h}^{i}_{\textit{[MASK]}} \in \mathbb{R}^d$. 

During training, the MLP classifier, $\mathcal{C}$, is optimized with the following objective function:

\begin{equation}
    \mathcal{J_{\text{CE}}}= -\frac{1}{N}\sum^{N}_{i=1}\textbf{y}^ilog(\mathcal{C}(\textbf{h}^{i}_{\textit{[MASK]}}))
\end{equation}
where $N$ is the number of training instances, $\textbf{y}^i$ is the one-hot encoding of the label $y^i$, and $\mathcal{C}(\cdot)$ is the network output of $\mathcal{C}$.

\subsection{Prompt-based Data Augmentation}
\label{subsec:data-augmentation}

As discussed in \S\ref{subsec:task-formulation}, the size of the training and development sets are small. If we learn $\mathcal{C}$ with just $\mathcal{D}^T_{train}$ and $\mathcal{D}^T_{dev}$, it is difficult for it to generalize to the unseen $\mathcal{D}^T_{test}$ where $|\mathcal{D}^T_{test}| \gg |\mathcal{D}^T_{train}| = |\mathcal{D}^T_{dev}|$. Therefore, we propose to augment $\mathcal{D}^T_{train}$ with an auxiliary language model. Note that the number of training parameters of the auxiliary language model is much smaller than that of $\mathcal{M}$. 

Motivated by the findings in previous works~\cite{gao-etal-2021-making,chen-etal-2021-revisiting} that prompt-based finetuning of the language model with demonstrations can drastically outperform standard fine-tuning procedures in the low resource setting, we apply prompt-based finetuning for learning a teacher model (the auxiliary language model), $\mathcal{A}$, which is then used to pseudo-label unlabeled text data. A filter mechanism is implemented to exclude pseudo-labeled data that the teacher model is less confident about.

\subsubsection{Prompt-based Finetuning With Demonstration}

\begin{figure}[h]
  \centering
  \includegraphics[width=\linewidth]{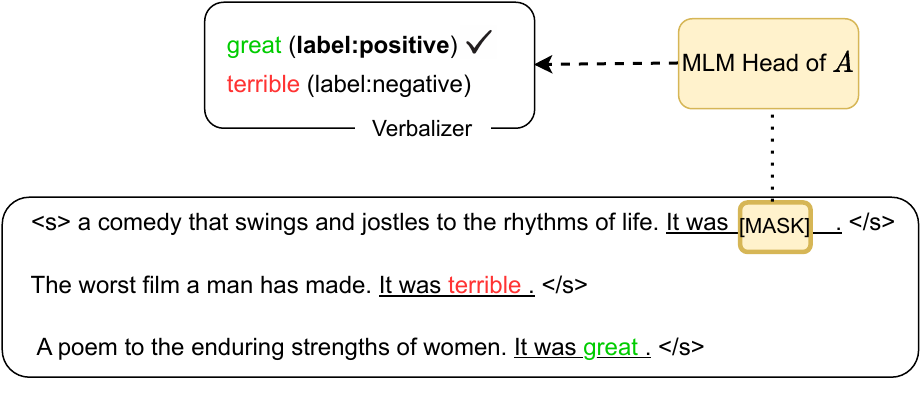}
  \caption{Prompt-based finetuning. The classification task is formulated as a cloze task where $\mathcal{A}$ learns to fill in the ``[MASK]" position. The underlined text is the prompt template. In the bottom box, the first line is the input text sequence. The second line is the demonstration for label:negative. The third line is the demonstration for label:positive. The verbalizer maps the labels to the corresponding words.}
  \label{fig:prompt-based learning}
\end{figure}

In this step, $\mathcal{D}^T_{train}$ and $\mathcal{D}^T_{dev}$ are used for training $\mathcal{A}$ and hyperparameter search, respectively. Figure~\ref{fig:prompt-based learning} illustrates the prompt-based finetuning process. Given $(X^i, y^i) \in \mathcal{D}^T_{train}$, the $X^i$ is first transformed into $\mathcal{F}_T(X^i)$ according to the task-specific templates\footnote{In our experiments, we use the same set of task-specific manual templates for both prompt-based finetuning of $\mathcal{A}$ and the training of $\mathcal{C}$.}. The verbalizer converts $y^i$ to the corresponding word in the vocabulary of $\mathcal{A}$. To fill in the ``[MASK]" position in $\mathcal{F}_T(X^i)$, $\mathcal{A}$ learns to assign a higher probability to the word mapped to $y^i$ than other label words. For example, $\mathcal{A}$ should predict a higher probability of ``great" than ``terrible" for the example input in Figure~\ref{fig:prompt-based learning}.

To further enhance the prompt-based finetuning process, we append demonstrations after $\mathcal{F}_T(X^i)$. A demonstration is an input text example. For instance, in Figure~\ref{fig:prompt-based learning}, ``The worst film a man has made. It was terrible" is a demonstration w.r.t. the negative class in the binary sentiment classification task. We append one demonstration for each label in the label space, $\mathcal{Y}$, to $\mathcal{F}_T(X^i)$.

$\mathcal{A}$ is finetuned with the standard MLM loss on $\mathcal{D}^T_{train}$. In addition, for model selection, we perform the grid search procedure on different training hyperparameters. The model variant with the best performance on $\mathcal{D}^T_{dev}$ is kept as the teacher model.

\subsubsection{Pseudo Labeling and Data Filtering}

The teacher model $\mathcal{A}$ is used to pseudo-label a large number of unlabeled text data. Specifically, an unlabeled text sequence is converted into the task-specific input format. $\mathcal{A}$ predicts the probability of words that correspond to the labels to fill in the ``[MASK]" position. The word (label) with the highest probability is the pseudo label of the text sequence. Motivated by previous works on pseudo-labeling~\cite{jacob-etal-2020-self,xie-etal-2020-uda,zhang-etal-2022-mdd}, we only keep pseudo-labeled text samples that $\mathcal{A}$ is confident about, i.e., the probability $\mathcal{A}$ assigned to the pseudo label is more than 90 percent. 

The unlabeled data are in-distribution w.r.t. task-specific few-shot training and development data. For each label of task $T$, we augment roughly the same amount of text data to ensure class-balanced training of $\mathcal{C}$. We denote the augmented set as  $\mathcal{D}^T_{aug}$. $\mathcal{C}$ is trained on $\mathcal{D}^T_{aug}\cup\mathcal{D}^T_{train}$ and evaluated on $\mathcal{D}^T_{dev}$. The model variant of $\mathcal{C}$ that performs the best on $\mathcal{D}^T_{dev}$ is selected for a final assessment on $\mathcal{D}^T_{test}$.

\section{Experiment}

\subsection{Preliminaries}
\label{subsec:prelim}


\noindent\textbf{Datasets} BT-Classifier is evaluated on 8 standard text classification datasets under a few-shot setting, including 4 single-sentence and 4 sentence-pair classification datasets. They are summarized in Table~\ref{tab:eval-data}. 

\begin{table}[!h]
\centering
    \caption{Statistics of the datasets. The ``single" column refers to whether the task is a single-sentence classification task. ``NLI" refers to natural language inference.}
\resizebox{\linewidth}{!}{
    \begin{tabular}{l|cccccc}
    \toprule
    Task Name & Single & \#Classes & Task Type & \#Train/Dev/Test & \#Augmented Data \\
    \midrule
    TREC~\cite{hovy-etal-2001-toward} & Yes & 6 & Question Type & 96/96/500 & $\sim$4.6K  \\
    AGNews~\cite{zhang-etal-2015-character} & Yes & 4 & Topic & 64/64/7.6K & $\sim$8.9K  \\
    Yelp~\cite{zhang-etal-2015-character} & Yes & 2 & Sentiment & 32/32/38K & $\sim$8.9K  \\
    SST-2~\cite{wang2018glue} & Yes & 2 &  Sentiment & 32/32/872 & $\sim$4K  \\
    MRPC~\cite{wang2018glue} & No & 2 & Paraphrase  & 32/32/1,725 & $\sim$3.1K  \\
    QQP~\cite{wang2018glue} & No & 2 & Paraphrase & 32/32/40.43K & $\sim$3K  \\
    QNLI~\cite{wang2018glue} & No & 2 & NLI  & 32/32/5,463 & $\sim$3K  \\
    SNLI~\cite{bowman-etal-2015-large} & No & 3 & NLI & 32/32/10K & $\sim$6K \\
    \bottomrule
    \end{tabular}
}
    \label{tab:eval-data}
\end{table}
\noindent  We set $K = 16$ for all the tasks. Following previous work, The training and dev splits of each task are randomly sampled from the original training set. Five independent sets of train and dev splits are sampled based on different random seeds. BT-classifier runs five times with these sets of data. Average performance (accuracy \%) on the original test set is reported for each task. If the original test set is not available, we evaluate BT-Classifier on the original dev set. The unlabeled data for augmentation are sampled from the original training set of each task, but with their original labels removed.

\bigskip
\noindent\textbf{Reproducibility} We adopt RoBERTa-Large~\cite{liu2019roberta} as the large-scale black-box language model. RoBERTa-Large consists of 24 transformer layers and the hidden size is 1024. In total, it contains 354 million parameters. We fix the architecture of MLP to be the same for all tasks, which is a 2-layer MLP with the Tanh activation function. For the teacher model, $\mathcal{A}$, we adopt DeBERTa-Base~\cite{he2021deberta}, which consists of 12 transformer layers and 100 million parameters. The hidden size of DeBERTa-Base is 768. Note that our approach is model-agnostic. 
This means that the black-box LLMs can be any encoder-only or encoder-decoder models and up to billions of parameters.
Moreover, the auxiliary teacher model can be any small encoder-only language model that can be finetuned with a reasonable amount of computational resources.

All the experiments are conducted on a single 24GB GeForce RTX 3090 GPU card. For learning teacher model $\mathcal{A}$, we set the training batch size, the maximum sequence length, and the maximum number of training steps as 2, 128, and 2000 respectively. We perform the grid search on the learning rate of (1e-5, 2e-5) and gradient accumulation steps (1, 2) respectively. For training the classifier $\mathcal{C}$, we set the train batch size, the total number of training epochs, and the maximum sequence length as 32, 100, and 512 respectively. The model is evaluated at the end of each epoch and if the validation accuracy doesn't improve for consecutive 5 epochs, we early stop the training process. Lastly, Table~\ref{tab:template} describes the label-word mapping and prompt templates we use in the experiments.

\begin{table}[!h]
\centering
    \caption{Task-specific prompt templates and label words.}
\resizebox{\linewidth}{!}{
    \begin{tabular}{l|ll}
    \toprule
    Task Name & Template & Label-Word Mapping \\
    \midrule
    TREC~\cite{hovy-etal-2001-toward} & [MASK] question: $<$X$>$ & direct use of the labels \\
    AGNews~\cite{zhang-etal-2015-character} & [MASK] News: $<$X$>$ & direct use of the labels \\
    Yelp~\cite{zhang-etal-2015-character} & $<$X$>$ . It was [MASK] . & negative: bad; positive: great \\
    SST-2~\cite{wang2018glue} & $<$X$>$ . It was [MASK] . & negative: bad; positive: great \\
    MRPC~\cite{wang2018glue} & $<$$X_1$$>$ ? [MASK] , $<$$X_2$$>$ & not\_Equivalent: no; equivalent: yes \\
    QQP~\cite{wang2018glue} & $<$$X_1$$>$ ? [MASK] , $<$$X_2$$>$ & not\_Equivalent: no; equivalent: yes \\
    QNLI~\cite{wang2018glue} & $<$$X_1$$>$ ? [MASK] , $<$$X_2$$>$ & entailment: yes; not\_entailment: no \\
    SNLI~\cite{bowman-etal-2015-large} & $<$$X_1$$>$ ? [MASK] , $<$$X_2$$>$ & entailment: yes; neutral: maybe; contradiction: no \\
    \bottomrule
    \end{tabular}
}
    \label{tab:template}
\end{table}

\begin{table*}[!t]
\centering
    \caption{Main experiment results. All results (accuracy \%) are the average across 5 different splits (\S\ref{subsec:prelim}), with which we perform the zero-shot evaluation. The standard deviation is reported in the bracket. $\dagger$ refers to white-box methods while $\ddagger$ refers to black-box methods. In the black-box category, the best score for each task is highlighted in bold and the second best is underlined. }
\resizebox{0.95\linewidth}{!}{
    \begin{tabular}{l|cccccccc|c}
    \toprule
     & TREC & AGNews & Yelp & SST-2 & MRPC &  QQP & QNLI & SNLI & Average \\
    \midrule
    Finetuning$\dagger$  & 88.8 (2.1) & 86.2 (1.4) & 91.8 (0.8) & 81.4 (3.8) &   76.6 (2.5) &  60.7 (4.3) & 56.3 (1.5) &  47.8 (6.8) &  76.2 \\
    LM-BFF$\dagger$ & 83.4 (2.7) &  87.1 (1.2) & 91.3 (2.7) & 92.3 (1.5) &  77.8 (2.0) &   69.8 (1.8) & 64.4 (4.6) & 76.5 (2.6) & 80.3  \\ \midrule
    ICL-RoBERTa$\ddagger$ & 26.2 (2.4) & 62.2 (13.5) & 85.4 (4.0) & 85.9 (0.7)  & 45.8 (6.7) & 36.1 (5.2)  & 53.8 (0.4) & 47.1 (0.6) & 53.0  \\
    Feature MLP$\ddagger$ & 25.3 (2.4) & 74.1 (2.0) & 79.2 (2.3) & 84.9 (3.8) &  68.4 (0.9) & \underline{64.8} (2.9) & 54.4 (4.5) & \underline{57.8} (3.2) & 63.6 \\
    BBT$\ddagger$ & 39.3 (5.2) & 81.2 (2.7) & \underline{91.5} (0.2) & 88.2 (1.7) & 61.6 (4.3) &  48.6 (8.3) & 56.8 (2.0) & 44.7 (4.0) & 65.8 \\
    BBTv2$\ddagger$ & 42.0 (4.5) & \underline{85.3} (0.5) & \textbf{92.9} (0.6) & \textbf{90.3}  (1.7) & \textbf{77.0} (4.7) & 56.3 (3.9)  & \underline{66.3} (2.3) & 57.3 (2.3) & \underline{70.9} \\ \midrule
    BT-Classifier (ours)$\ddagger$ & \textbf{78.4} (5.6) & \textbf{86.1} (1.0) & \underline{91.5} (2.9) & \underline{88.5} (2.9) & \underline{75.3} (4.9) & \textbf{77.8} (2.9)  & \textbf{66.6} (0.7) & \textbf{63.4} (1.0) & \textbf{80.3}  \\
    \bottomrule
    \end{tabular}
}
    \label{tab:exp-results}
\end{table*}

\bigskip
\noindent\textbf{Baselines} We compare BT-classifier with full-model fine-tuning methods and state-of-the-art black-box tuning methods described as follows: (1) \textbf{Finetuning}, the standard way of finetuning a language model for few-shot text classification. (2) prompt-based fine-tuning as implemented by Gao et al. (2021)~\cite{gao-etal-2021-making}. The approach is referred to as \textbf{LM-BFF}. Both (1) and (2) require updating the weights of the LLM. Hence, they can be seen as white-box methods. (3) \textbf{Feature MLP}, which is equivalent to BT-Classifier without prompt-based data augmentation. (4) \textbf{ICL-RoBERTa}, which applies the in-context learning approach proposed in Brown et al. (2020)~\cite{brown-etal-2020-language}. (5) \textbf{Black-Box Tuning (BBT)}~\cite{sun-etal-2022-bbt}. (6) \textbf{BBTv2}~\cite{sun-etal-2022-bbtv2}. (5) and (6) are derivative-free optimization methods that are based on the covariance matrix adaptation evolution strategy~\cite{hansen-ostermeier-2001-completely}. All the baselines use RoBERTa-Large as the backbone.

\subsection{Results \& Analysis}

\noindent\textbf{Main Analysis} The main results are summarized in Table~\ref{tab:exp-results}. Overall, we can observe BT-Classifier achieves an average accuracy score of 80.3\% over the 8 different classification tasks, outperforming the second-best black-box method, BBTv2 by 5.5\% absolute scores. For each task, BT-Classifier is either the best or the second best in the black-box method category. Additionally, BT-classifier outperforms standard finetuning and achieves comparable results to LM-BFF, the prompt-based finetuning method in terms of the average accuracy across the eight tasks. 

Specifically, BT-Classifier performs much better than other black-box method on TREC, which has 6 categories. BT-Classifer also performs stably well on the more challenging NLI tasks while BBTv2 doesn't. In addition, the observation that Feature MLP performs much worse than BT-Classifier justifies the effectiveness of data augmentation for improving model generalization. The significant performance gap between LM-BFF and standard finetuning justifies our adoption of prompt-based finetuning for learning the auxiliary teacher model. Hence, with the help of the prompt-based finetuned teacher, the MLP on top of the LLM fully utilizes the thousands of augmented data and outperforms other black-box approaches. 

\bigskip
\noindent\textbf{Ablation Study} we analyze the performance of different BT-Classifier variants. Table~\ref{tab:ablation-results} summarizes the results.

\begin{table}[!h]
\centering
    \caption{Ablation results of BT-Classifier.}
\resizebox{0.9\linewidth}{!}{
    \begin{tabular}{l|ccc|c}
    \toprule
      & Teacher & CLS Token & Last Layer & BT-Classifier \\ \midrule
     TREC & 63.8 (3.8) & 80.7 (3.5) & 81.4 (3.7) & 78.4 (5.6) \\
     AGNews & 84.7 (0.7) &  84.9 (1.8) & 85.2 (4.1) & 86.1 (1.0) \\
     Yelp & 87.9 (2.2) & 91.8 (2.0) & 92.2 (2.3) & 91.5 (2.9) \\
     SST-2 & 82.5 (4.5) & 84.4 (5.9) & 89.7 (3.2) & 88.5 (2.9) \\
     MRPC & 64.3(5.3) & 77.2 (2.6) & 74.3 (5.5) & 75.3 (4.9) \\
     QQP & 68.8 (2.6) & 75.9 (5.0) & 74.7 (2.9) & 77.8 (2.9) \\
     QNLI & 60.8 (3.7)  & 62.0 (3.1) & 62.7 (2.2) & 66.6 (0.7) \\
     SNLI & 62.0 (5.2) & 59.7 (1.2) & 61.7 (2.1) & 63.4 (1.0)  \\ \midrule
     Average & 71.9 & 77.1 & 77.7 & 80.3 \\ 
    \bottomrule
    \end{tabular}
}
    \label{tab:ablation-results}
\end{table}
\noindent The ``Teacher" column contains the performance of the finetuned auxiliary teacher model on different tasks. We can observe that on average, BT-Classifier outperforms the teacher model by an absolute accuracy score of 8.4\%. This demonstrates that BT-Classifier is more robust and generalizes better to unseen test sets than the teacher model even though the pseudo labels of the augmented data are imperfect. 


Furthermore, we analyze the effect of extracting the hidden states w.r.t. the start token instead of those of the ``[MASK]" token. The results are presented in the column ``CLS Token". A performance drop of 3.2\% on average is observed. This observation proves that the hidden states w.r.t. the ``[MASK]" position carry more indicative information for text classification under the prompt-based setting.

Lastly, as shown in the ``Last Layer" column, where the hidden state from the last layer of the black box is utilized,  the average accuracy score drops by 2.6\% than that of BT-Classifier. The observation aligns with findings in previous works on sentence representation learning~\cite{zhang-etal-2020-unsupervised,zhang-etal-2021-bootstrapped,wang-etal-2020-sbert} that hidden states from multiple transformer layers carry richer information than the hidden state from just the final transformer layer.

\bigskip
\noindent\textbf{Efficiency Analysis} The total number of tunable parameters of BT-Classifier is 1.05M, which is much less than that of RoBERTa-Large (354M). Hence, our approach is much more parameter-efficient than full-model tuning methods, which include standard finetuning and LM-BFF. 
 Furthermore, compared to BBT or BBTv2, the training process of BT-Classifier is much faster. For example, for the AGNews task, BBT requires around 88 min to complete training while the training time of BT-Classifier is around 37 min. Note that the training times of BBT and BT-Classifier are computed based on learning of 64 training samples and 8.9K training samples respectively. 

Even though compared to feature MLP, BT-Classifier requires additional time for data augmentation and a longer training period to learn the additional augmented text data, the performance gain of BT-Classifier over feature MLP is significant, which is 12.8\% absolute accuracy score on average (Table~\ref{tab:exp-results}). Additionally, the data augmentation process also doesn't take too much time. The grid search of the teacher model is roughly 20 minutes (4 model variants * 5 min per model). The inference of 8K unlabeled text data takes roughly 1 minute.

Hence, BT-Classifier can be an excellent choice of parameter-efficient adaptation of LLMs under computational resource constraints. 

\section{Conclusion \& Future Work}

In summary, we propose an efficient and effective approach, BT-Classifier, for black-box few-shot text classification. Our proposed method achieves state-of-the-art performance among different parameter-efficient approaches for black-box LLM adaptation. It also achieves comparable results to methods that require full-model tuning of the LLMs. Two major reasons contribute to the superior performance of BT-Classifier: (1) data augmentation with prompt-based finetuning and (2) the strong text semantic representation of different transformer layers of the feature extractor, i.e., the black-box LLM. One limitation of BT-Classifier is that it requires abundant unlabeled in-domain text for data augmentation. Such in-domain text may not be necessarily available in practical scenarios. Hence, future work can explore how to leverage the general-domain text for data augmentation. In addition, BT-Classifier can also be applied for parameter-efficient adaptation of large-scale speech pre-trained language models on spoken language understanding tasks.

\bibliographystyle{IEEEtran}

\end{document}